\documentclass[letterpaper, 10 pt, conference]{ieeeconf}  

\IEEEoverridecommandlockouts                              

\usepackage[table, dvipsnames]{xcolor}
\usepackage{amsmath} 
\usepackage{graphicx}
\usepackage{adjustbox}
\usepackage{multicol}
\usepackage{tabularx}
\usepackage{caption}
\usepackage{amssymb}
\usepackage{cellspace}
\usepackage{hyperref}
\usepackage{cite}
\usepackage{multirow}
\usepackage{float}
\usepackage{booktabs}
\usepackage{subcaption}
\usepackage{soul}
\let\labelindent\relax
\usepackage{enumitem}
\usepackage{xspace}
\usepackage{bm}

\newcommand{\stat}[2]{\ensuremath{#1_{#2}}}
\newcommand{\coolname}{DefFlow\xspace}

\newcommand{\best}[1]{%
  \cellcolor{blue!10}{\boldmath #1}%
}
\title{\LARGE \bf
\coolname: Goal-Conditioned Deformable Object Manipulation using 3D Point Flows
}

\author{%
Gaurav Singh, Sergio Orozco, Ryan Duong, Mete Tuluhan
Akbulut, George Konidaris, and Srinath Sridhar\\
{\normalsize Brown University, Providence, RI, USA}%
}

\begin{document}

\makeatletter
\def\@IEEEauthorblockconfadjspace{-1em}

\let\@oldmaketitle\@maketitle
\renewcommand{\@maketitle}{%
    \@oldmaketitle
    \centering
    \vspace{0pt}
    \includegraphics[width=\linewidth]{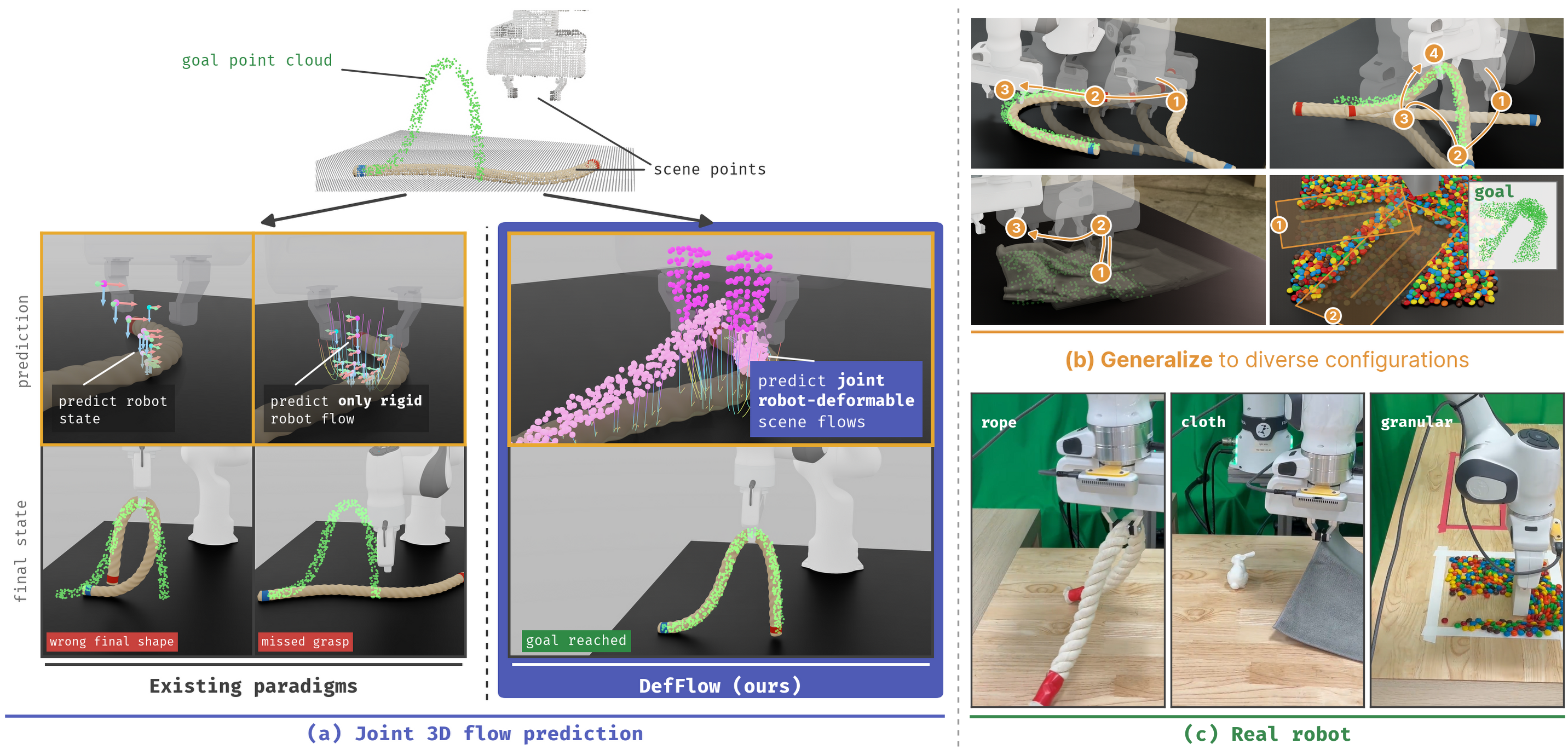}
    \captionof{figure}{
    \small \textbf{\coolname learns to manipulate deformable objects by predicting goal-conditioned joint robot-scene point flows.} This enables it to learn goal-directed manipulation behaviors across diverse 3D configurations.}
    \label{fig:teaser}
}
\makeatother

\maketitle
\setcounter{figure}{1}
 \thispagestyle{empty}
\pagestyle{empty}


\begin{abstract}

Manipulating deformable objects toward desired 3D configurations is a fundamental capability for robots operating in unstructured physical environments. 
However, as the range of possible initial and target configurations grows, goal-conditioned deformable manipulation requires a robot to infer increasingly diverse manipulation behaviors in addition to precise continuous actions.
We introduce \coolname, a framework for goal-conditioned manipulation of deformable objects toward a broad range of 3D configurations. \coolname learns from demonstrations by jointly predicting the goal-conditioned 3D point flow of the robot and deformable object. 
We show that this formulation better captures both the manipulation decisions, such as when and where to grasp, release, or re-grasp, and the precise continuous motions required for goal reaching, rather than predicting only the robot flow.
Across rope, granular-media, and cloth manipulation settings with diverse initial and target configurations, \coolname substantially outperforms goal-conditioned action-prediction and robot-flow-only baselines on success rates and goal-distance metrics. 
We further demonstrate deployment of our framework on a real robot using real point-cloud observations and unseen goal configurations. Project Page:
\textcolor{blue}{\href{https://def-flow.github.io}{https://def-flow.github.io}}.

\end{abstract}


\section{INTRODUCTION}

Deformable-object manipulation often involves reaching a desired target shape or configuration, such as routing cables or ropes into curved shapes, wrapping or draping cloth, or shaping granular material. 
Enabling robots to achieve a diverse range of such goals is central to developing more general and versatile deformable-object manipulation systems~\cite{yin2021modeling}.

However, developing goal-conditioned policies that can reliably reach a broad range of such goal configurations remains challenging for several reasons.
\textbf{First}, unlike rigid objects, whose configurations can be described by 6-DoF poses, deformable objects can undergo a wide range of 3D shape changes governed by complex dynamics.
\textbf{Second}, reaching a specific goal requires modeling the coupled dynamics of the robot and deformable object, capturing how their configurations evolve together toward the target~\cite{ arriola2020modeling}.
\textbf{Finally}, as the space of possible initial and target configurations becomes broader, the behaviors required to connect them become increasingly diverse, and can be completely different for different goals or starting configurations even for the same object within the same setting. 
For example, reaching one rope configuration may require a single grasp and drag, while another may require twisting, releasing, and re-grasping at a different location. 

Prior works have explored learning goal-conditioned deformable object policies from 3D pointclouds~\cite{deformernet,bartsch2024sculptdiff,bartsch2025pinchbot}.
However, these approaches typically show settings with relatively narrower distributions of object configurations and isolated behavior primitives, such as a single grasping action followed by servoing~\cite{deformernet} or only pinching/squeezing behaviors~\cite{bartsch2024sculptdiff,bartsch2025pinchbot}.
Using dense 3D flows as the underlying representation has been shown to improve policy learning from pointclouds for deformable tasks~\cite{toolflownet}, however it is limited to only predicting rigid tool flows~\cite{toolflownet}.

In this work, we introduce \textbf{\coolname}, a goal-conditioned policy for manipulating deformable objects toward a broad range of 3D configurations using dense 3D point flows. 
\textit{Our \textbf{key insight} is to use \textbf{goal-conditioned joint robot-deformable object 3D point flow} as the underlying representation learning objective and repurpose that as the robot policy using the robot flows.}
This formulation, combined with our data curation recipe enables better goal-conditioned representation learning by associating robot behaviors with their corresponding object deformations, helping distinguish behaviors through their effects on the scene and generalizing to a diverse set of goal shapes that require completely different action trajectories and decisions to reach.
We find that compared to only predicting the robot motion~\cite{toolflownet}, the goal-conditioned 3D flows of the full scene together can improve goal-reaching performance and better predict the manipulation behaviors such as grasping, releasing, and re-grasping, together with the precise continuous actions that the robot must execute to make the objects reach diverse goal configurations. 

This recipe has been explored in 2D image space~\cite{ye2026wam, vera}: models that learn the coupled scene prediction in the same output space, e.g. video models, can be repurposed as better policies as the robot motion predicted maps to better action priors. 
This has been explored mostly for settings with defined tasks (usually using language conditioning)~\cite{ye2026wam, vera}.
For goal-conditioned manipulation, while a similar recipe has been explored with 3D pointclouds, it is limited to rigid objects whose motion is predicted using 6-DoF poses~\cite{lyu2025dywa}. In contrast, we explore this recipe in goal-conditioned deformable manipulation using dense 3D point flow prediction as the underlying objective.
We test \coolname on a large number of varied configurations across rope, granular-media, and cloth manipulation with broad start and goal distributions, showcasing the ability of our approach to generalize to diverse configurations and outperform baselines that predict only robot actions or robot flow.

To summarize, our contributions are:
\begin{itemize}
\item We introduce \coolname, a goal-conditioned policy for manipulating deformable objects toward a broad range of 3D configurations requiring diverse manipulation behaviors.

\item We show that learning to predict the goal-conditioned 3D flows for the full scene in addition to robot-flow improves generalization to diverse goal-shapes, decisions and continuous actions for goal reaching.

\item We show the effectiveness of our approach in various complex goal-reaching environments with a large number of target configurations. We also further validate our approach by deploying it on a real robot with unseen goal shapes.

\end{itemize}

\section{Related Works}
\label{sec:rw}
\vspace{-3pt}

\subsection{3D Point Flows for Manipulation}
\vspace{-2pt}
3D point flow is a popular representation for modeling 3D dynamics~\cite{toolflownet, huang2026pointworld, eisner2022flowbot3d}. They are also called particle-based representations, where each particle is an entity moving through the 3D space~\cite{zhang2025particle, zhang2024adaptigraph, huang2025particleformer}. Previous works have mostly used 3D pointflows for representing forward dynamics, i.e. for building action-conditioned world models or transition functions that take the current state and the robot actions as input and predict the future state evolution conditioned on the actions. While transition functions can be used for model-predictive control, selecting actions for goal reaching remains a nontrivial planning problem. Existing approaches simplify this problem through task-specific action restrictions, such as starting from pre-grasped configurations~\cite{huang2026pointworld}, using tabletop pushing or grasp-and-drag actions~\cite{zhang2024adaptigraph}, or parameterizing gripping and lifting behaviors~\cite{li2018learning,shi2024robocraft,huang2025particleformer}.

Goal-conditioned policies directly predict continuous robot actions from the current observation and goal (Sec.~\ref{sec:rw_gc}) and have been explored using 2D flows~\cite{weng2022fabricflownet}. Closer to our approach, ToolFlowNet~\cite{toolflownet} addresses the challenge of policy learning from point clouds for deformable manipulation by predicting dense 3D flows. However, it imposes an SVD-based rigidity constraint on the flows and thus predicts flow only on rigid tools/end-effectors rather than on the deformable object. Our approach doesn't impose this rigid constraint and directly predicts the full-scene 3D flow conditioned on a goal configuration.

\vspace{-2pt}
\subsection{Goal-Conditioned Deformable Object Manipulation}
\label{sec:rw_gc}
\vspace{-3pt}

Goal-conditioned deformable manipulation has been studied across tasks including cable rearrangement, fabric folding, shape servoing, and deformable sculpting~\cite{seita2021learning,deformernet,bartsch2024sculptdiff,bartsch2025pinchbot, weng2022fabricflownet}. 
Early works commonly formulated manipulation in a top-down 2.5D setting, where a policy selects top-down planar grasp and placement positions, with a fixed grasping height~\cite{seita2021learning}.
More recent works consider goal-conditioned manipulation directly with 3D pointclouds~\cite{deformernet,bartsch2024sculptdiff,bartsch2025pinchbot}. 
~\cite{deformernet} considers varied goal shapes, but limits the configurations considered to those reachable through a single grasp followed by lifting, usually done with anchored tissue in surgical retraction.
Other works~\cite{bartsch2024sculptdiff,bartsch2025pinchbot} have applied 3D diffusion policies~\cite{ze20243d} to goal-conditioned deformable manipulation, demonstrating deformable object shaping through structured squeezing or pinching behaviors.
However, these approaches have not been successfully demonstrated in settings with diverse 3D configurations, where the challenge is selecting both the correct contact mode (e.g., whether to grasp or release) and the precise 3D actions required (i.e., where to grasp and which motion trajectory to follow).
Moreover, these approaches do not use the 3D flows as a dense representation objective.

\begin{figure}[t]
\centering
\includegraphics[width=\linewidth]{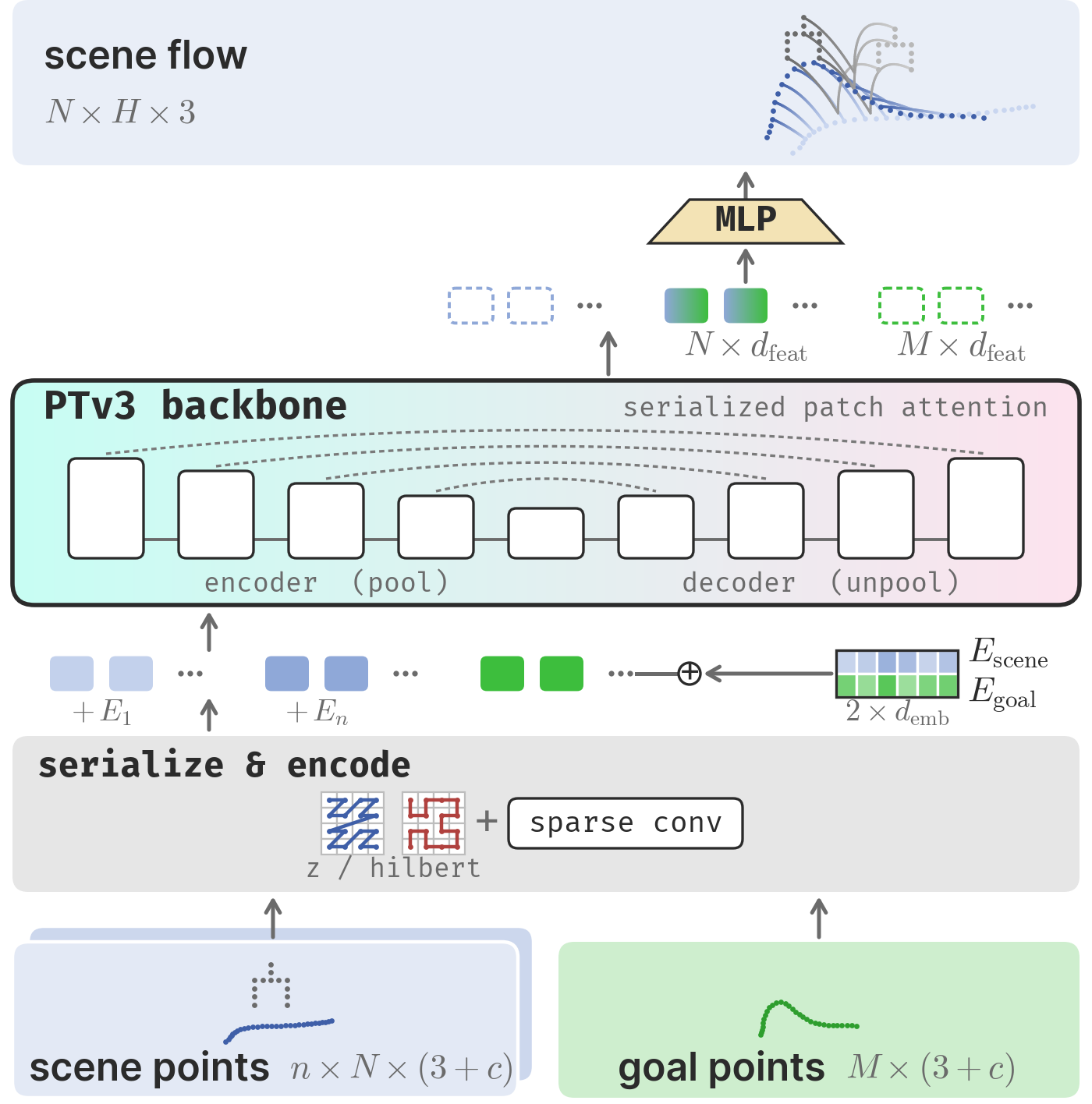}
\caption{\small \textbf{\coolname architecture.} Given the scene (including the robot) and a goal point cloud, \coolname processes all points in a unified 3D space using a PTv3 backbone with learned type-embeddings, which is passed through an MLP to get 3D scene flow. The robot flows are then used to recover executable robot actions.
}
\label{fig:arch}
\vspace{-10pt}
\end{figure}

\section{\coolname: Goal-Conditioned Policy using 3D Point Flow Prediction}
\label{sec:method}

\subsection{Problem Formulation}

We consider the problem of learning 3D goal-conditioned manipulation of deformable objects from pointclouds. At time $t$, the robot receives an observation history of the robot and scene, together with a goal $g$ specifying the desired 3D configuration of the deformable object. Let $o_{\leq t}$ denote the observation history and $a_t$ the robot action. The objective is to learn a policy $\pi_\theta(a_t \mid o_{\leq t}, g)$ whose actions drive the object from its initial configuration toward the goal.

We assume access to a fixed dataset of goal-conditioned demonstrations $\mathcal{D}=\{(\tau_i,g_i)\}_{i=1}^{X}$, where $\tau_i=(o_0^i,a_0^i,\ldots,o_{T_i-1}^i,a_{T_i-1}^i,o_{T_i}^i)$ records the observations and actions of a demonstration of length $T_i$, and $g_i$ is its associated goal. 

Let $P^s_t\in \mathbb{R}^{N\times 3}$ and $P^g\in \mathbb{R}^{M\times 3}$ be the scene and goal points respectively at time $t$.
Here $P^s_t$ crucially includes points on the robot as well.
Given the observation history $P^s_{\leq t}$ and the goal point cloud, \coolname predicts future scene point flows over a prediction horizon of $H$ timesteps as
\begin{equation}
    \Delta P^s_{t+1:t+H}
    =
    f_\theta(P^s_{\leq t}, P^g).
\end{equation}

\begin{figure*}[t]
\centering
\includegraphics[width=\linewidth]{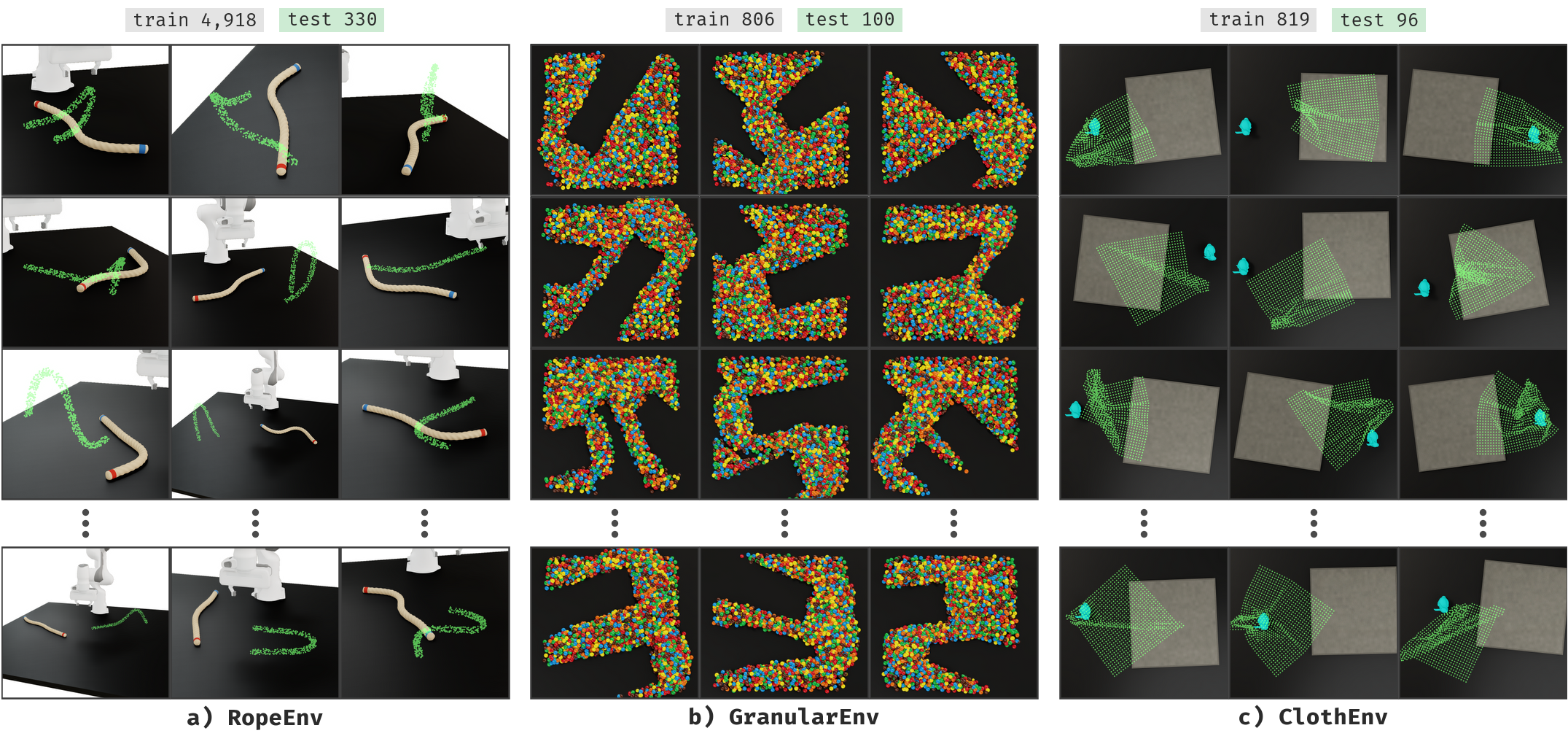}
\caption{\small Simulation benchmarking environments. This figure shows the distribution of tasks in each environment. As can be seen in (a), each task differs not only in the deformed shape, orientation, and position of both start and goal configurations, but also in the action trajectory required to reach the goal. Some require grasping the middle of the rope, others an end or a point in between. Some require only one grasp, while others require multiple grasps; some may require dragging forward, to the left, or in a zigzag pattern, or swishing. For (b), each task involves a different target shape and requires a different combination of paddle sweeps, varying in number, angle, and extent. Some target shapes require the sweeps to be performed in a specific order. In (c), tasks differ not only in the cloth's starting position and orientation and its goal configuration, but also significantly in the actions required, as some might require grasping the middle of an edge and simply draping the cloth over the bunny, while others require a corner grasp, a slight swish, and then draping. All three environments share large scene configuration spaces and combinatorial action spaces, much broader than those explored previously~\cite{bartsch2024sculptdiff, seita2021learning, bartsch2025pinchbot, toolflownet}.} 
\label{fig:configs}
\vspace{-10pt}
\end{figure*}

\textbf{Pipeline.}
We build our modeling representations based on 3D pointflows similar to previous approaches such as PointWorld and ToolFlowNet~\cite{toolflownet,huang2026pointworld}, which adapt point-cloud UNet-style backbones to obtain per-point features for task-specific action prediction and action-conditioned world prediction through per-point flows. Here we outline our recipe for adapting a pointcloud UNet backbone for goal-conditioned scene-flow prediction.
We use Point Transformer V3 (PTv3)~\cite{wu2024point} as the UNet backbone of our flow prediction model.
Our recipe is outlined in Fig.~\ref{fig:arch} and is as follows.
Each point is represented by its 3D coordinate together with optional channels for color.
Given $\mathbf{P}^{s}_{t-n+1:t}$ and $\mathbf{P}^{g}$, where n is the number of observations, we concatenate the points and pass them to the backbone as a single unified set, followed by serialization and sparse-convolution encoding to obtain per-point tokens $(N\times n+M) \times d$. We add learnable embedding vectors $\mathbf{E}_{s},\mathbf{E}_{g}\in \mathbb{R}^d$ to the scene and goal tokens respectively, and an additional learnable temporal positional embedding $E_i\in \mathbb{R}^d, i \in [1,n]$ for observations.
Since the goal and the rest of the scene are in the same coordinate frame, this provides a simple way to input points from multiple sources to the same PTv3 backbone without requiring a separate cross-attention module~\cite{wu2024point}.
The output is a per-point feature representation for all points.
A small per-point MLP maps each robot and scene feature at the current timestep to its future displacements.
This produces the joint robot and scene flow $\Delta \mathbf{P}^{s}_{t+1:t+H}$ defined above.
To obtain per-point correspondences for our experiments, we sample points on the scene geometry as well as the robot URDF and keep the point identities fixed across timesteps.
For our specific settings, we limit the robot points to those sampled on the gripper for simplicity, although our formulation supports including points on any number of links of the robot URDF.

Given the ground-truth scene flows from the demonstration, we train using a Huber loss averaged over points:
\begin{equation}
\label{eq:flow_loss}
\mathcal{L}_{\mathrm{flow}}
=
\sum_{i=1}^{H}
\mathcal{L}_{\mathrm{Huber}}
\left(
\Delta \mathbf{P}^{s}_{t+i},
\Delta \hat{\mathbf{P}}^{s}_{t+i}
\right).
\end{equation}

\subsection{Action Extraction via Inverse Kinematics.}
To recover executable actions from points, we solve for the robot joint configurations whose forward kinematics reproduce these point positions. Let $N_r$ denote the number of sampled robot points. For each future timestep $t+h$, with $h \in \{1,\ldots,H\}$, we solve $\mathbf{q}^{*}_{t+h} = \arg\min_{\mathbf{q}} \sum_{i=1}^{N_r} \|\mathbf{P}^{r}_{\mathrm{FK},i}(\mathbf{q}) - \hat{\mathbf{P}}^{r}_{t+h,i}\|_2^2$ using damped least-squares inverse kinematics, where $\mathbf{P}^{r}_{\mathrm{FK}}(\mathbf{q}),\hat{\mathbf{P}}^{r}_{t+h}$ are the robot point clouds computed from the URDF at joint configuration $\mathbf{q}$ (including the gripper finger prismatic joints), and robot pointflows respectively. The resulting action sequence can be sent to existing robot controllers, allowing \coolname to produce executable actions without a separate learned action head.

\section{Experiments}
\label{sec:experiments}
To evaluate the behavior priors learned by \coolname for goal reaching, we conduct quantitative experiments across multiple challenging deformable-object simulation settings.
Our benchmarks are designed to span different interaction modes, deformation dynamics, and levels of behavior-sequencing complexity. In this section, we describe our benchmark environments, demonstration collection procedure, evaluation protocol and metrics.

\textbf{Simulation Benchmark and Data Curation.}
Motivated by representative settings studied in prior deformable-manipulation literature~\cite{bartsch2024sculptdiff, seita2021learning, bartsch2025pinchbot, zhang2024adaptigraph, zhang2025particle}, we evaluate on three domains that span distinct object dynamics and interaction patterns: (a) rope/cable manipulation, (b) granular-media shaping, and (c) cloth manipulation. We construct three simulation benchmark environments using Newton~\cite{newton} and Warp-based spring-mass simulation frameworks~\cite{warp2022, zhang2025real, jiang2025phystwin}. Each environment isolates a different source of difficulty:

\texttt{\textbf{1. RopeEnv}} focuses on multi-stage prehensile manipulation of a rope. The robot can grasp, drag, lift, twist, and release the rope at different locations, and compose these behaviors in varying orders with continuously varying motion magnitudes, e.g., rotating the rope by $45^\circ$ or translating it by $20\,\mathrm{cm}$. We use the physics engine from~\cite{zhang2025real, jiang2025phystwin}, which supports rigid--deformable contact between the robot fingers and rope. This is the most challenging environment out of the three, testing the ability to execute multi-stage behavior sequences from the current and goal configurations, including transitions between grasping, manipulation, release, and re-grasping.

\texttt{\textbf{2. GranularEnv}} focuses on non-prehensile interaction with a granular sand pile. The robot is equipped with a paddle tool that can sweep through the material in varying directions and over varying distances, with the number and sequence of sweeps also allowed to vary. Unlike rope and cloth, the granular material can be reshaped and sculpted into substantially different shapes through repeated applications of the same basic sweeping primitive. This environment tests whether the policy can infer the direction, magnitude, and sequence of continuous non-prehensile interactions required to reach a target shape. We use the Material Point Method (MPM) implementation in Newton~\cite{newton} to simulate the granular material.

\texttt{\textbf{3. ClothEnv}} focuses on prehensile cloth manipulation involving interaction with a rigid obstacle. The robot can grasp the cloth along its edge, drag it around a rigid object in varying directions and over varying distances, and subsequently release it. The resulting cloth configuration depends not only on the robot trajectory but also on contact with the rigid object as the cloth bends, wraps, and slides around it. This environment tests whether the policy can account for complex contact-induced deformations as the cloth bends, wraps, and slides around the rigid object while inferring the manipulation behavior required to reach a target configuration. We use the Vertex Block Descent (VBD) implementation in Newton~\cite{newton} to simulate the cloth.

Our simulation environments are shown in Fig~\ref{fig:configs}. 
Within each environment, the initial and target configurations vary substantially, producing a diverse distribution of start--goal task instances, as illustrated in Fig~\ref{fig:configs}. 
Different start--goal pairs may require qualitatively different sequences of manipulation behaviors, or substantially different continuous instantiations of the same interaction primitive through changes in direction, magnitude, duration, and ordering.

\textit{Collecting Demonstrations.} Obtaining expert demonstrations for diverse deformable-object goals in a goal-first manner is non-trivial. 
We therefore avoid assuming access to such a goal-conditioned expert and instead collect unstructured play data in simulation, covering diverse object configurations and manipulation behaviors~\cite{lynch2020learning, yin2026playworld}. For each environment, we construct procedural policies that use privileged state information to compose interaction primitives such as grasping, pushing, pick-and-place, and randomized motions from randomized initial configurations. Following hindsight relabeling~\cite{andrychowicz2017hindsight}, we treat the final state of the deformable object in the episode as its goal.
A key nuance is that this procedural method is not the same as sampling from a set of fixed tasks. It is a composition of continuous primitives, with different numbers of them in each episode. This gives us a dataset of significantly different behaviors and unseen goals as shown in Fig~\ref{fig:configs}. Thus each demonstration is closer to its own ``task'', giving us a dataset of arbitrary tasks, tied to their goal state and starting state, and the held out task contains settings without any direct counterpart in the training data, both in terms of the 3D shapes and behaviors.
Across the three environments, we generate a total of nearly 7,000 randomized tasks, with the allocation weighted by the diversity of each environment's configuration and interaction space. Accordingly, \texttt{RopeEnv} contains the largest share, while \texttt{GranularEnv} and \texttt{ClothEnv} use fewer demos due to their more constrained interaction settings.

\begin{figure}
    \centering
    \includegraphics[width=\linewidth]{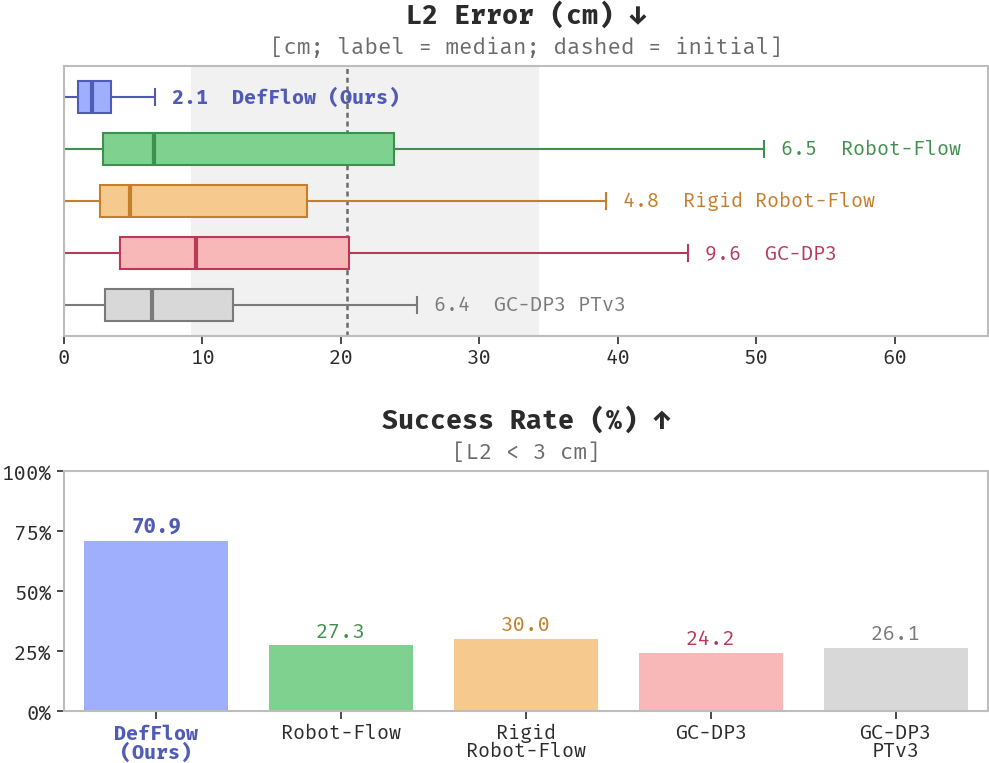}
    \caption{Quantitative results on \texttt{RopeEnv}. Lower is better for $L_2$ error, higher for SR.}
    \label{fig:ropemetrics}
\end{figure}

\begin{table}[t]
\centering
\setlength{\tabcolsep}{3pt}

\resizebox{\columnwidth}{!}{%
\begin{tabular}{@{}lccccc@{}}
\toprule
& \multicolumn{3}{c}{\texttt{GranularEnv}}
& \multicolumn{2}{c}{\texttt{ClothEnv}} \\
\cmidrule(lr){2-4}
\cmidrule(lr){5-6}
\textbf{Method}
& \textbf{CD}$\downarrow$
& \textbf{EMD}$\downarrow$
& \textbf{3D IoU}$\uparrow$
& \textbf{$L_2$ (cm)}$\downarrow$
& \textbf{Succ@3cm}$\uparrow$ \\
\midrule

GC-DP3
& \stat{1.05}{0.18}
& \stat{2.41}{0.54}
& \stat{55.66}{7.05} \
& \stat{6.60}{6.10}
& 25\% \\

GC-DP3-PTv3
& \stat{0.99}{0.18}
& \stat{2.27}{0.79}
& \stat{58.44}{7.23} \
& \stat{6.58}{7.05}
& 22.9\% \\

Robot-Flow
& \stat{0.95}{0.15}
& \stat{2.02}{0.54}
& \stat{57.80}{7.03} \
& \stat{6.06}{6.81}
& 32.3\% \\

Rigid Robot-Flow
& \stat{0.92}{0.15}
& \stat{1.90}{0.45}
& \stat{59.00}{7.76} \
& \stat{6.02}{6.79}
& 32.3\% \\

GC-PointBERT
& \stat{0.92}{0.17}
& \stat{2.07}{0.57}
& \stat{58.78}{7.76} \
& --
& -- \\

Grasp-Servo
& --
& --
& --
& \stat{4.77}{3.77}
& 33.3\% \\

\textbf{\coolname (Ours)}
& \best{\stat{0.82}{0.18}}
& \best{\stat{1.76}{0.64}}
& \best{\stat{64.14}{8.27}} \
&  \best{\stat{3.33}{1.95}}
& \best{\textbf{49.0\%}} \\

\bottomrule
\end{tabular}%
}
\caption{Quantitative results on \texttt{GranularEnv} and
\texttt{ClothEnv}. CD, EMD, and $L_2$ are reported in cm, with means and standard deviations over episodes;
3D IoU is shown $\times 10^2$.}
\label{tab:main_results}
\vspace{-15pt}
\end{table}

\begin{figure*}[t]
\centering
\includegraphics[width=\linewidth]{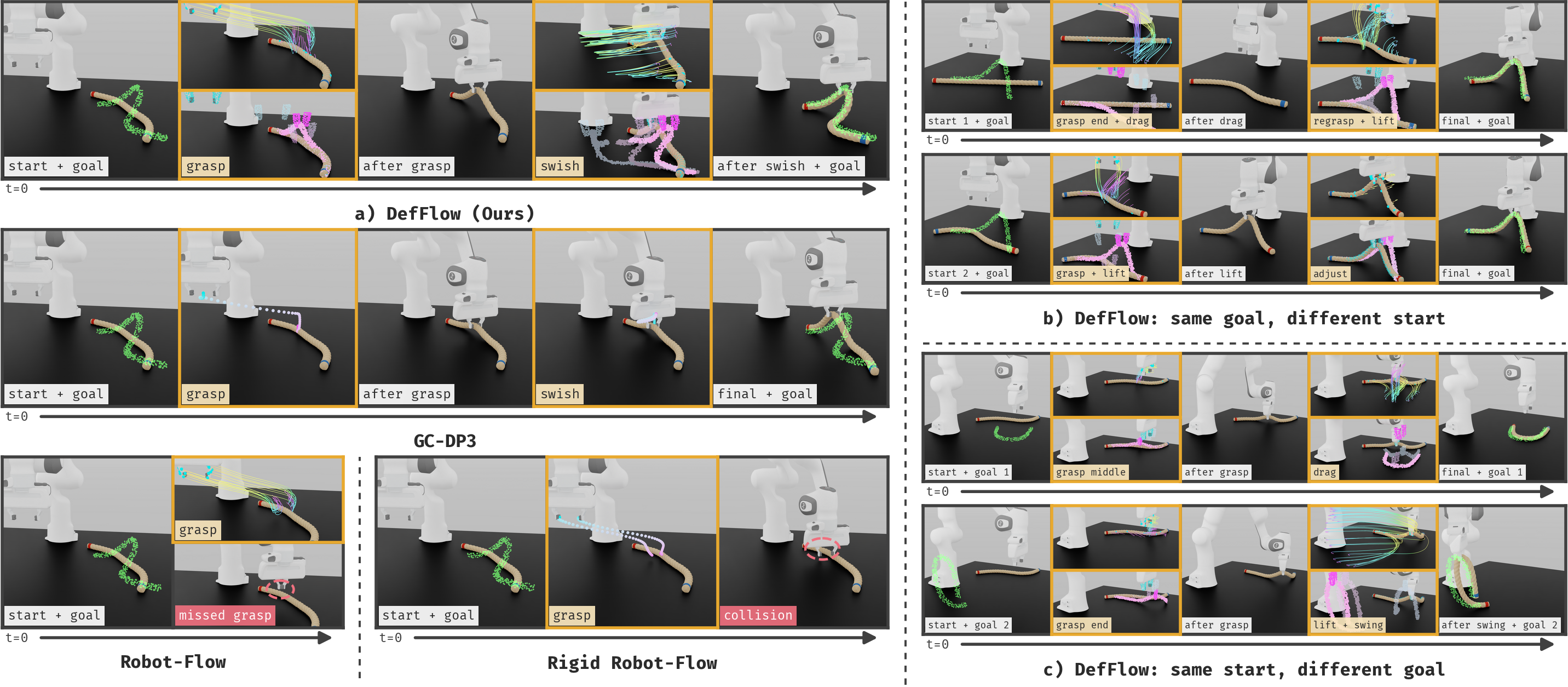}
\caption{\small Qualitative Results on rope manipulation. In (a) the robot needs to grasp the rope, lift and swish back in a precise manner to reach the exact shape. As can be seen, \coolname predicts the precise grasp+swishing, while the baselines either get the grasp incorrect and collide with the rope, or do not predict the precise swish resulting in the rope not reaching the exact shape. In (b) we showcase two settings with the same goal but different initial configurations, one from the same demo as the goal and the other from a separate demo, while in (c) we show settings with the same starting state but another goal from a separate demo. As can be seen, \coolname also generalizes to these unseen settings by predicting different action trajectories as in (c) and taking different decisions in (b) where for one it grasps and lifts while for the second it first grasps and moves the rope to align it and then re-grasps and lifts.
}
\label{fig:rope_qualitative}
\vspace{-10pt}
\end{figure*}
\begin{figure}[t]
\centering
\includegraphics[width=\linewidth]{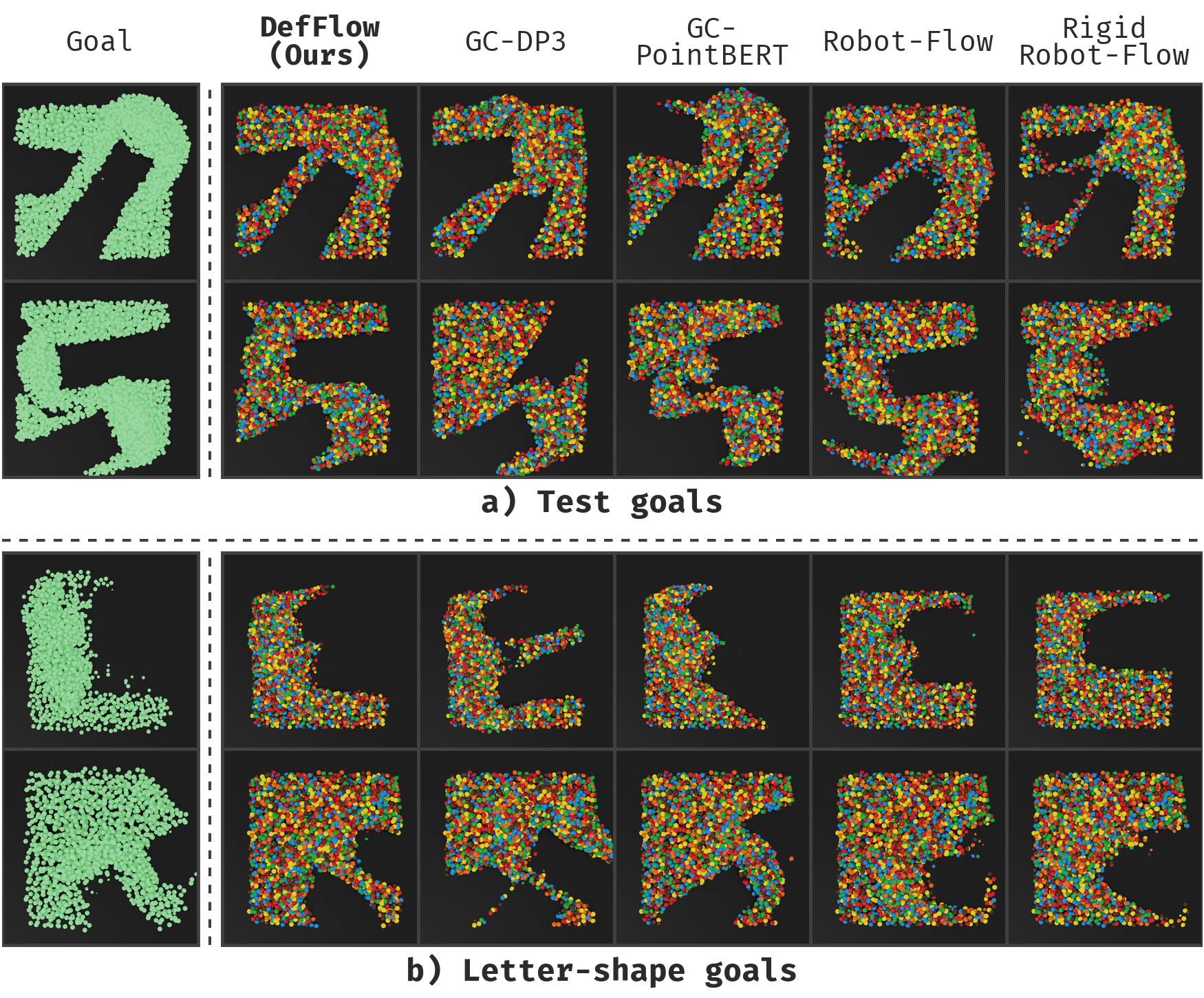}
\caption{\small Qualitative Results on Granular Shaping. We show qualitative results on unseen goal shapes in (a). We additionally show results on unseen letter shapes in (b). Each goal shape is its own unique design requiring different sweep actions to accomplish in different orders and numbers.  
}
\label{fig:gran_qualitative}
\vspace{-20pt}
\end{figure}

\textbf{Baselines.}
Prior deformable-manipulation methods are evaluated in substantially different experimental setups, including different objects, robot embodiments, sensing configurations, simulators, and task definitions, making direct end-to-end comparisons across a common benchmark difficult to interpret~\cite{bartsch2024sculptdiff, bartsch2025pinchbot, toolflownet, deformernet}. For a clean comparison, we structure our baselines into two types: (a) Ablations, where we ablate over our key design decision (robot-deformable scene flow) covering existing closest prior work~\cite{toolflownet} and (b) Adaptations of other goal-conditioned deformable manipulation works, where we adapt each family of prior work using their codebases to preserve the key ideas and design decisions while holding the training data, goal distributions, evaluation environments and input/output hyperparameters fixed for all.~\footnote{We also attempted building a baseline with MPC + PointWorld~\cite{huang2026pointworld}. However, an MPC pipeline has not been successfully shown in prior work on deformable object tasks as complex as ours and with multiple behaviors to even search on top of and chain them, and we were unable to get it to converge on any plausible results. While the forward model worked, it was not useful as the MPC action sampling itself was the major bottleneck.}

\textbf{1. Ablations.} The key design decision difference between ToolFlowNet and \coolname is the prediction of deformable object flow, 
while the architectures are in the same family. We structure this comparison as two ablations: (a) \textbf{Robot-Flow Only,} which uses the same architecture, inputs, and training data as \coolname, but removes supervision on the deformable-scene flow and trains only on robot-flow prediction; and (b) \textbf{Rigid Robot-Flow (SVD)}, where the predicted
robot/gripper point flows of each finger are projected to a rigid $SE(3)$ motion
using a differentiable SVD-based alignment layer from ToolFlowNet.

\textbf{2. Baselines from other prior works:} 
We also compare with other non-flow based approaches. We adapt the following baselines to our evaluation setting: (a) GC-DP3: We adapt 3D Diffusion Policy~\cite{ze20243d} to the goal-conditioned setting by conditioning the policy on both the current scene observation and target point cloud. This baseline represents the family of approaches that directly predict
goal-conditioned robot actions, including diffusion-policy-based
deformable manipulation methods such as SculptDiff~\cite{bartsch2024sculptdiff} and PinchBot~\cite{bartsch2025pinchbot}. We evaluate two encoder variants:
\textbf{GC-DP3} uses the original 3D Diffusion Policy MLP-based point-cloud
encoder, while \textbf{GC-DP3-PTv3} uses a PTv3 encoder. 

For \texttt{GranularEnv}, the granular pile remains centered in the workspace, making the object-centric representation used by SculptDiff~\cite{bartsch2024sculptdiff} directly applicable. Thus we also add a task-specific baseline \textbf{GC-PointBERT} with the pretrained PointBERT as the encoder. For \texttt{ClothEnv}, we additionally evaluate a structured goal-conditioned policy that first predicts a grasp point and then predicts the post-grasp robot motion, adapted from DeformerNet~\cite{deformernet}, called \textbf{Grasp-Servo}. This factorization captures the core assumption of prior structured deformable-manipulation approaches that explicitly prescribe the interaction structure, including grasp-selection followed by shape servoing or placement \cite{deformernet,seita2021learning}. Note that this is not a direct fair comparison as the action primitives for that task are provided to the baseline whereas all other baselines and our approach have to learn that behavior from data.

We do not impose an analogous structured-policy baseline on
\texttt{RopeEnv}, since its evaluation distribution explicitly contains
variable-length sequences of grasping, dragging, lifting, twisting,
releasing, and re-grasping. Prescribing a fixed grasp--motion structure
would therefore exclude a substantial portion of the benchmark by
construction. Likewise, \texttt{GranularEnv} already operates through a
single structured interaction family of paddle sweeps, whose direction,
magnitude, number, and ordering must be inferred from the goal; an
additional hand-designed primitive factorization would therefore largely
reparameterize the existing action space rather than provide a distinct
comparison.

\textbf{Metrics.}
We evaluate goal reaching using metrics appropriate to the underlying
object representation. For connected deformable objects
(\texttt{RopeEnv} and \texttt{ClothEnv}), simulator particles maintain
consistent correspondence and ordering throughout an episode. We
therefore measure the mean per-particle Euclidean distance between the
final configuration and the target configuration. We additionally report
\textbf{Succ@3cm}, where an episode is considered successful if this
final-state mean per-particle $L_2$ distance is below $3\,\mathrm{cm}$.
For \texttt{GranularEnv}, particle-wise correspondence is not a
meaningful measure of the achieved pile shape, since individual particles
can undergo substantial rearrangement while producing similar overall
geometry. We therefore use correspondence-free set- and shape-level
metrics: Chamfer distance (CD), Earth Mover's Distance (EMD), and voxelized 3D Intersection-over-Union (IoU). CD and EMD measure geometric discrepancy between the final and
target point sets, while 3D IoU measures volumetric overlap between their
voxelized shapes.

\section{Results and Ablations}
\label{sec:results}

\textbf{Qualitative results.}
Figures~\ref{fig:rope_qualitative},~\ref{fig:gran_qualitative} showcase the qualitative results of our approach. As can be seen, \coolname predicts accurate joint scene evolution showcasing the rope being swished and deformed, and the precise coupled action together, and thus the executed actions accurately reach the goal state. For granular, \coolname predicts the precise sweeps along with the deformation of the pile with that sweep. This coupling between the robot motion and deformable object allows it to distinguish between very similar actions during training (e.g. swishes of different extents and directions, or sweeps that have very similar action trajectories) as the coupled deformation for them is significantly different which leads it to predict more accurate robot motion during inference as shown. As can be seen \coolname reaches the granular designs, whereas the baselines often sweep incorrectly either leaving out parts or over-sweeping resulting in suboptimal designs. Please check Supp. Video for additional results.

\textbf{Quantitative results.}
Table~\ref{tab:main_results} and Fig~\ref{fig:ropemetrics} summarize performance across the three deformable-manipulation environments. As can be seen, \coolname outperforms action-only baselines as well as robot-only flow baselines across all the three environments.
For \texttt{RopeEnv}, \coolname greatly improves the success rate as well as the final goal shape L2 error, being $>2\times$ better than the next best baseline on L2 and Succ@3cm. This environment is the most difficult of them as it has the largest configuration space where the rope can be in arbitrary 3D shapes, curves, on the ground, lifted etc in the final shape. For \texttt{GranularEnv}, \coolname gets the best IoU compared to other baselines. For \texttt{ClothEnv}, \coolname achieves a $1.5\times$ higher success rate than the strongest baseline.
The Robot-Flow and ToolFlowNet-style ablations reach a similar performance compared to other action-only baselines.
A surprising effect we find is that even though \coolname performs better than existing baselines, the improvement in performance is more for environments with larger configuration space and with more varied initial and goal configurations. 
Although all the environments have highly diverse configuration spaces, the relative ordering between them from high to low is (a) rope, varying positions, orientations and shapes, (b) granular with less position variation (as the pile is centered) but more shape variation, and (c) cloth, where the demonstrations broadly start around the obstacle and end nearer to it. 
While it is not trivial to control the configuration spaces of such robot-deformable coupled systems easily especially for things like cloth where naively increasing random behaviors can just crumple the cloth into similar shapes, we probe more into the effect of configuration spaces by a more controlled setting as follows:

\begin{figure}[hbt!]
    \centering
    \includegraphics[width=\linewidth]{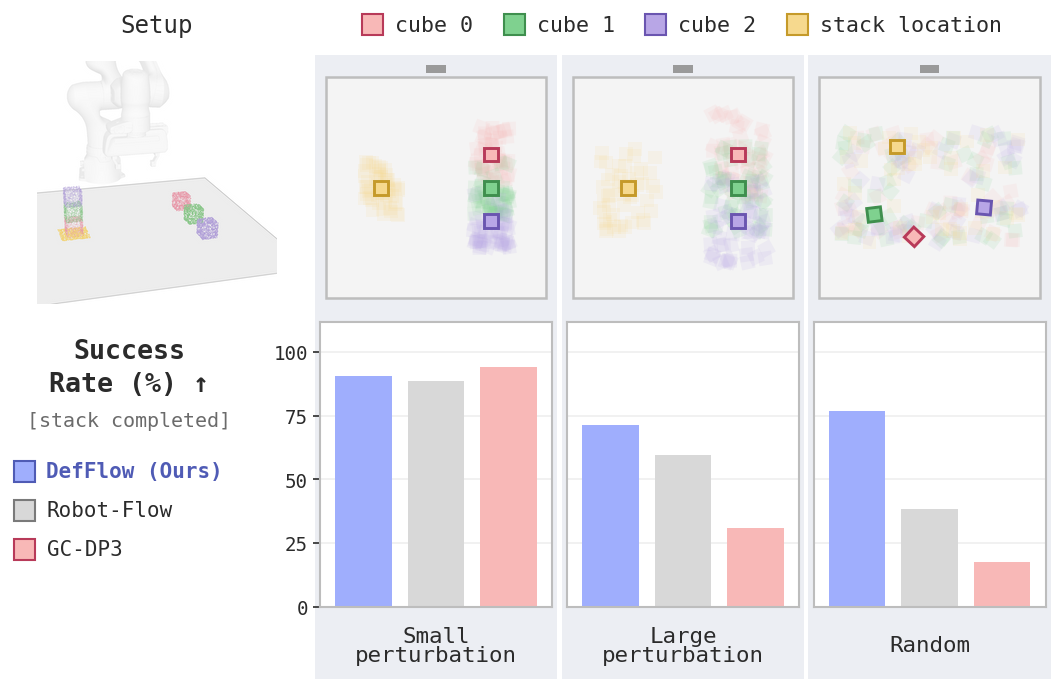}
    \caption{\small Toy cube-stacking experiment setting and results. The three columns show the top-down view of the distribution and the perturbations as well as the density of the demonstrations, with the corresponding success rate bar plots below them}
    \label{fig:cube}
\end{figure}

\textbf{Effect of Configuration Diversity.}
We use a simple toy cube-stacking setting to further study how performance changes
as the range of demonstrated configurations increases in a controlled manner. The setup is shown in Fig~\ref{fig:cube}. Three cubes are
initialized at varying positions and must be stacked, in a specified order,
at a target location. We progressively increase the randomization of the
cube poses and stack location, from less perturbed configurations to fully
randomized ones. We collect around 700 demos for each setting and keep 52 held out for evaluation. Importantly, the demonstrations in each setting do cover the task distribution of that setting but increasingly sparsely as the configuration space grows. Thus, a policy can no longer rely on repeatedly seeing similar trajectories and must instead generalize the underlying pick-and-place behavior across novel cube poses, locations, and target locations. As shown in Fig.~\ref{fig:cube}, while both GC-DP3 and Robot-Flow perform comparably in less perturbed settings, with dense demonstration coverage, their success rates drop quickly as the perturbation increases, whereas \coolname degrades more gracefully.
This suggests that, for tasks within a constrained distribution, the range of possible behaviors is narrow, thus making it easier to choose an action using just the start and end goal states. Here, coupled scene prediction possibly offers a small improvement or stays comparable. However, as configuration diversity increases, perhaps it becomes harder to reason about what to do without having the coupled association of how the robot-scene system evolves, and full scene flow prediction offers higher gains. This experiment sheds light on the effect of configuration diversity and offers a possible explanation for the differences in results across the three deformable environments. Note that the three deformable environments do not have the same amount of training data, and while we try our best to keep the number of demos somewhat proportional to the breadth of the action distribution, it is hard to control the deformable configuration distribution, unlike in the cube-stacking setting.

\textbf{Real-Robot Deployment.}
We demonstrate \coolname on a physical robot setup consisting of a Franka Emika Panda arm and an Intel RealSense D455 camera (Fig.~\ref{fig:teaser}(c)). We use the respective checkpoints trained on \texttt{RopeEnv}, \texttt{GranularEnv}, and \texttt{ClothEnv}. For rope manipulation, we showcase the pretrained general goal-conditioned policy used for a specific new task: \textit{``Hang a rope on the stand''}, simply by specifying the desired final rope configuration as the goal. Without any task-specific fine-tuning, we run the same policy across multiple initial rope configurations and two goals. 
For granular, we try to shape the pile into unseen letter shapes (L, C, and S). For cloth we try unseen goal configurations requiring a specific section of cloth to be draped over the object. 
We use a simple object estimation pipeline that first uses SAM2~\cite{ravi2025sam} to segment the object points. We perform object estimation for rope by using the segmented image to get the rope spline, and then use the partial pointcloud and the rope spline to get the full rope points. For granular material, we use the segmentation mask to identify the occupied positions on the table. For cloth, we directly use the segmented RGB-D point cloud, as the simulated cloth is represented by a grid of surface points.
As seen in the Supp. video, our method is sufficiently good to support real life results on unseen goal-shapes and it showcases how a policy trained over diverse start--goal configurations can be used for a new task by changing only the specified goal state.
Rollouts and additional details are provided in the Supp. video.

\section{Conclusion} 
\label{sec:conclusion} 
We introduce \coolname, a goal-conditioned policy for deformable-object manipulation over diverse 3D start and goal configurations, formulated through a joint 3D flow objective. As configuration diversity increases, goal reaching requires inferring not only precise continuous motions, but also the appropriate manipulation behaviors for a given start--goal pair. \coolname addresses this by jointly predicting robot and deformable-scene point flows, learning goal-dependent robot behavior together with the resulting scene evolution from demonstrations collected using procedural play data and hindsight goal relabeling. Across rope, granular-media, and cloth manipulation, our experiments show that joint robot--scene prediction improves goal-reaching performance, while a controlled toy study further demonstrates its advantage as the configuration space becomes broader and more sparsely demonstrated. We also demonstrate execution of a simulation-trained policy on a physical robot.

\bibliographystyle{IEEEtran}

\bibliography{bibtex}

\end{document}